\documentclass[letterpaper, 10 pt, conference]{ieeeconf}
\IEEEoverridecommandlockouts
\usepackage[utf8]{inputenc}
\usepackage[T1]{fontenc}
\usepackage[english]{babel}

\usepackage{amsmath}
\usepackage{amssymb}
\usepackage[ruled,vlined,linesnumbered]{algorithm2e}

\usepackage[
	activate   = {true},
	protrusion = false,
	expansion  = true,
	kerning    = true,
	spacing    = true,
	tracking   = false,
	auto       = true,
	selected   = true,
	factor     = 2000,
	stretch    = 30,
	shrink     = 40,
]{microtype}
\usepackage{stackengine}
\usepackage{graphicx}
\usepackage{soul}
\usepackage{tabularx}
\usepackage[nolist]{acronym}
\usepackage{mathtools}
\usepackage{amsfonts}
\usepackage{subcaption}
\usepackage{booktabs}
\usepackage[table]{xcolor}

\makeatletter
\let\NAT@parse\undefined
\makeatother
\usepackage[pdfa,colorlinks,bookmarksopen,bookmarksnumbered,allcolors=black]{hyperref}
\usepackage{cite}
\usepackage[nospread]{flushend}

\usepackage[nameinlink,capitalise]{cleveref}
\crefname{line}{line}{lines}
\crefname{figure}{Fig.}{Figs.}
\Crefname{figure}{Fig.}{Figs.}
\crefname{equation}{Eq.}{Eqs.}
\Crefname{equation}{Eq.}{Eqs.}
\crefname{section}{Sec.}{Secs.}
\Crefname{section}{Sec.}{Secs.}
\crefname{table}{Tbl.}{Tbls.}
\Crefname{table}{Tbl.}{Tbls.}
\crefname{definition}{Def.}{Defs.}
\Crefname{definition}{Def.}{Defs.}
\crefname{algorithm}{Alg.}{Algs.}
\Crefname{algorithm}{Alg.}{Algs.}
\crefname{line}{L.}{Ls.}
\Crefname{line}{L.}{Ls.}
\crefname{assumption}{Asm.}{Asms.}
\Crefname{assumption}{Asm.}{Asms.}
\crefname{subassumption}{Asm.}{Asms.}
\Crefname{subassumption}{Asm.}{Asms.}
\Crefname{problem}{Problem}{Problems}
\crefname{problem}{Problem}{Problems}

\DeclareMathOperator*{\argmin}{argmin}

\graphicspath{{../pdf/}{/figs/}}

\makeatletter
\newcommand\notsotiny{\@setfontsize\notsotiny{6.2}{7}}
\makeatother

\SetArgSty{textup}

\title{\LARGE \bf
    Nearest-Neighbourless Asymptotically Optimal Motion Planning with Fully Connected Informed Trees (FCIT*)

}

\author{Tyler S.\ Wilson$^{1}$, Wil Thomason$^{2}$, Zachary Kingston$^{2,3}$, Lydia E.\ Kavraki$^{2,4}$, and Jonathan D.\ Gammell$^{1}$%
\thanks{$^{1}$Estimation, Search, and Planning (ESP) Research Group, Queen's University, Kingston ON, Canada. \texttt{\{18tsw1,gammell\}@queensu.ca.}}%
\thanks{$^{2}$Department of Computer Science, Rice University, Houston TX, USA \texttt{\{wbthomason, zak, kavraki\}@rice.edu}}%
\thanks{$^{3}$Department of Computer Science, Purdue University, West Lafayette IN, USA}%
\thanks{$^{4}$Ken Kennedy Institute, Rice University, Houston TX, USA}%
\thanks{This work was supported by the Natural Sciences and Engineering Research Council of Canada (NSERC) [RGPIN-2024-06637], and the National Science Foundation (NSF) [2336612].}%
}

\newcommand{\squeezeWords}{\looseness=-1}

\SetKwIF{If}{ElseIf}{Else}{if}{}{else if}{else}{end if}%

\NewDocumentCommand{\tablecell}{m m}{
\notsotiny
\begin{tabular}{@{}c@{}}
	#1 \\
	#2
\end{tabular}
}

\definecolor{lightergray}{gray}{0.925}

\begin{document}

\maketitle
\thispagestyle{empty}
\pagestyle{empty}

\begin{acronym}
	\acro{VAMP}{Vector-Accelerated Motion Planning}
	\acro{OMPL}{Open Motion Planning Library}
	\acro{ASAO}{almost-surely asymptotically optimal}
	\acro{RRT}{Rapidly-exploring Random Trees}
	\acro{G-RRT*}{Greedy RRT*}
	\acro{SIMD}{single instruction/multiple data}
	\acro{BIT*}{Batch Informed Trees}
	\acro{PRM}{Probabilistic Roadmaps}
	\acro{ABIT*}{Advanced BIT*}
	\acro{GBIT*}{Greedy BIT*}
	\acro{FIT*}{Flexible Informed Trees}
	\acro{MBM}{MotionBenchMaker}
	\acro{RGG}{random geometric graph}
	\acro{PEA*}{Partial Expansion A*}
	\acro{EPEA*}{Enhanced PEA*}
	
	\acro{FCIT*}{Fully Connected Informed Trees}
    \acro{DoF}{degree-of-freedom}
\end{acronym}

\begin{abstract}

Improving the performance of motion planning algorithms for high-\acl{DoF} robots usually requires reducing the cost or frequency of computationally expensive operations.
Traditionally, and especially for asymptotically optimal sampling-based motion planners, the most expensive operations are local motion validation and querying the nearest neighbours of a configuration.

Recent advances have significantly reduced the cost of motion validation by using \ac{SIMD} parallelism to improve solution times for satisficing motion planning problems.
These advances have not yet been applied to asymptotically optimal motion planning.

This paper presents \ac{FCIT*}, the first fully connected, informed, anytime \ac{ASAO} algorithm.
\ac{FCIT*} exploits the radically reduced cost of edge evaluation via \ac{SIMD} parallelism to build and search fully connected graphs.
This removes the need for nearest-neighbours structures, which are a dominant cost for many sampling-based motion planners, and
allows it to find initial solutions faster than state-of-the-art \ac{ASAO} (\acs{VAMP}, \acs{OMPL}) and satisficing (\acs{OMPL}) algorithms on the \acl{MBM} dataset while converging towards optimal plans in an anytime manner.
\squeezeWords

\end{abstract}

\acresetall

\section{Introduction}

Planning low-cost motions for high-\ac{DoF} robots quickly is a fundamental area of research in robotics.
These high-\acs{DoF} robots are described by a continuous \emph{configuration space}, but motion planning requires both a discrete approximation of this space and the ability to efficiently search this approximation.
Graph-based planners, such as Dijkstra's algorithm~\cite{dijkstra} and A*~\cite{astar}, require the configuration space (i.e., search space) to be discretized \textit{a priori}, and both their planning time and the quality of their solution depends on the resolution of this discretization.

Sampling-based motion planners, such as \ac{PRM}~\cite{prm}, \ac{RRT}~\cite{rrt}, and RRT-Connect~\cite{rrtc}, avoid this \textit{a priori} discretization by incrementally sampling the search space which constructs a \ac{RGG} online as a discrete approximation of this search space.
Anytime \ac{ASAO} sampling-based motion planners, such as RRT*~\cite{prmstar} and \ac{BIT*}~\cite{bit}, extend sampling-based motion planning by continually improving their sampled approximations even after finding an initial solution in order to converge probabilistically towards an optimal solution.
This search in \ac{BIT*} is ordered by potential solution cost, minimizing the required number of edge evaluations (i.e., local motion validations). %

\begin{figure}[t]
    \centering
    \begin{subfigure}[b]{\linewidth}
    	\includegraphics[width=.33\textwidth]{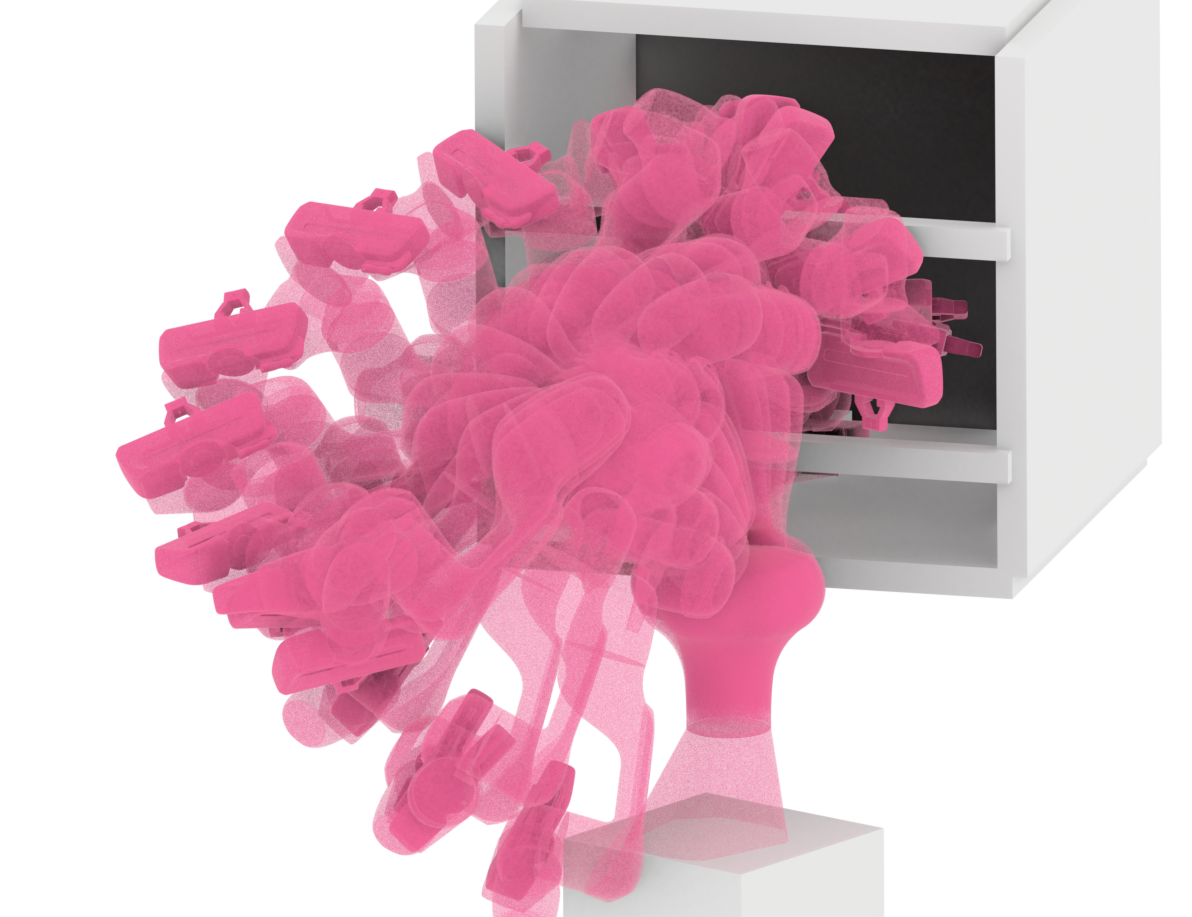}\hfill
        \includegraphics[width=.33\textwidth]{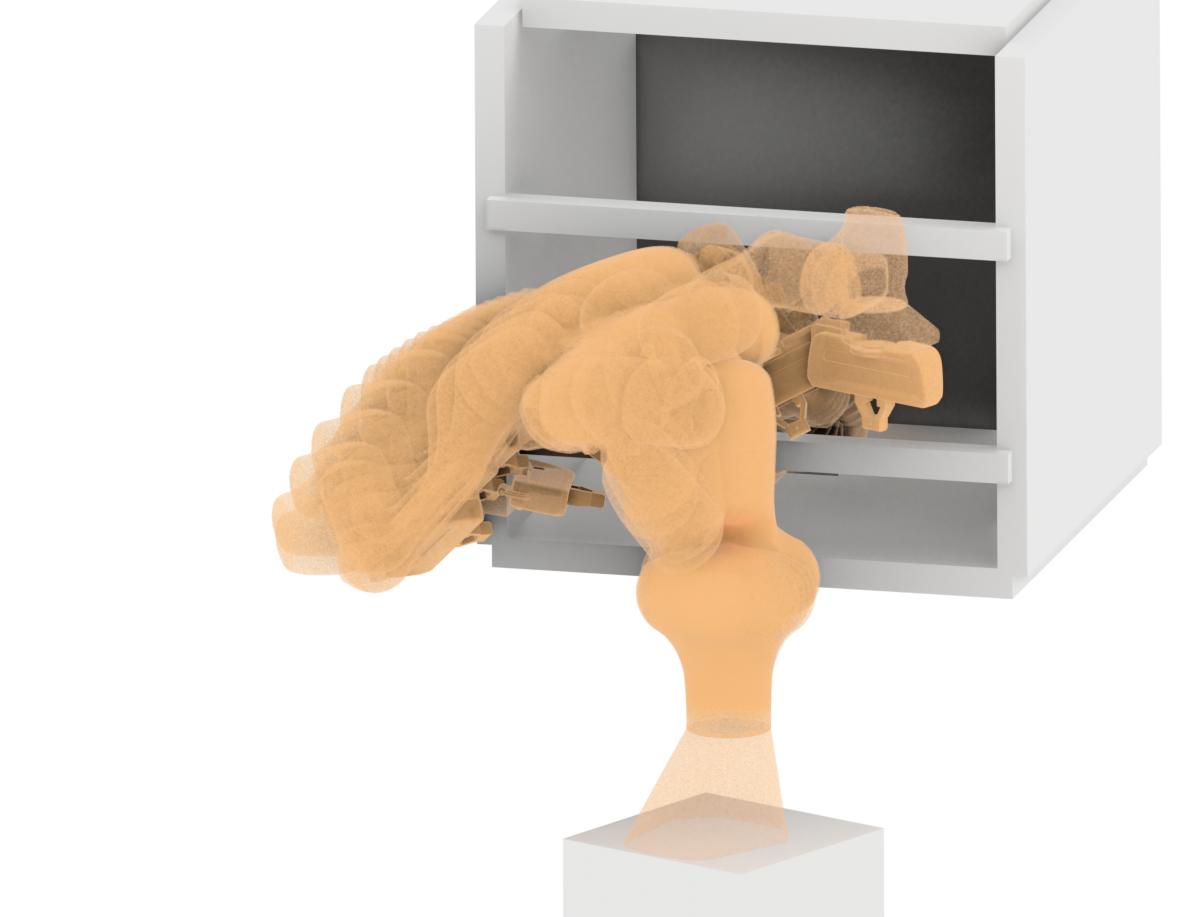}\hfill
        \includegraphics[width=.33\textwidth]{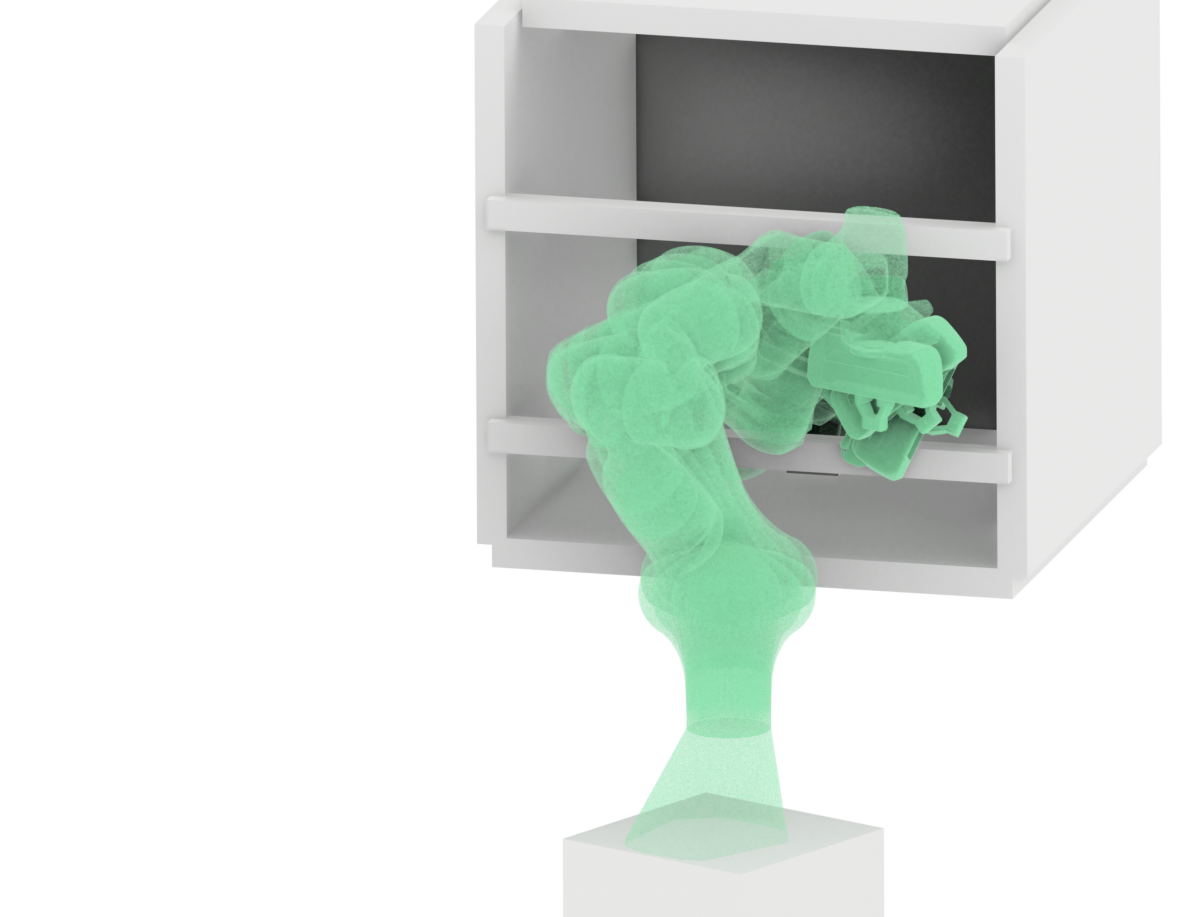}
        \caption{Computed Motions}
    \end{subfigure}

	\centering
    \begin{subfigure}[b]{\linewidth}
    	\includegraphics[width=\linewidth]{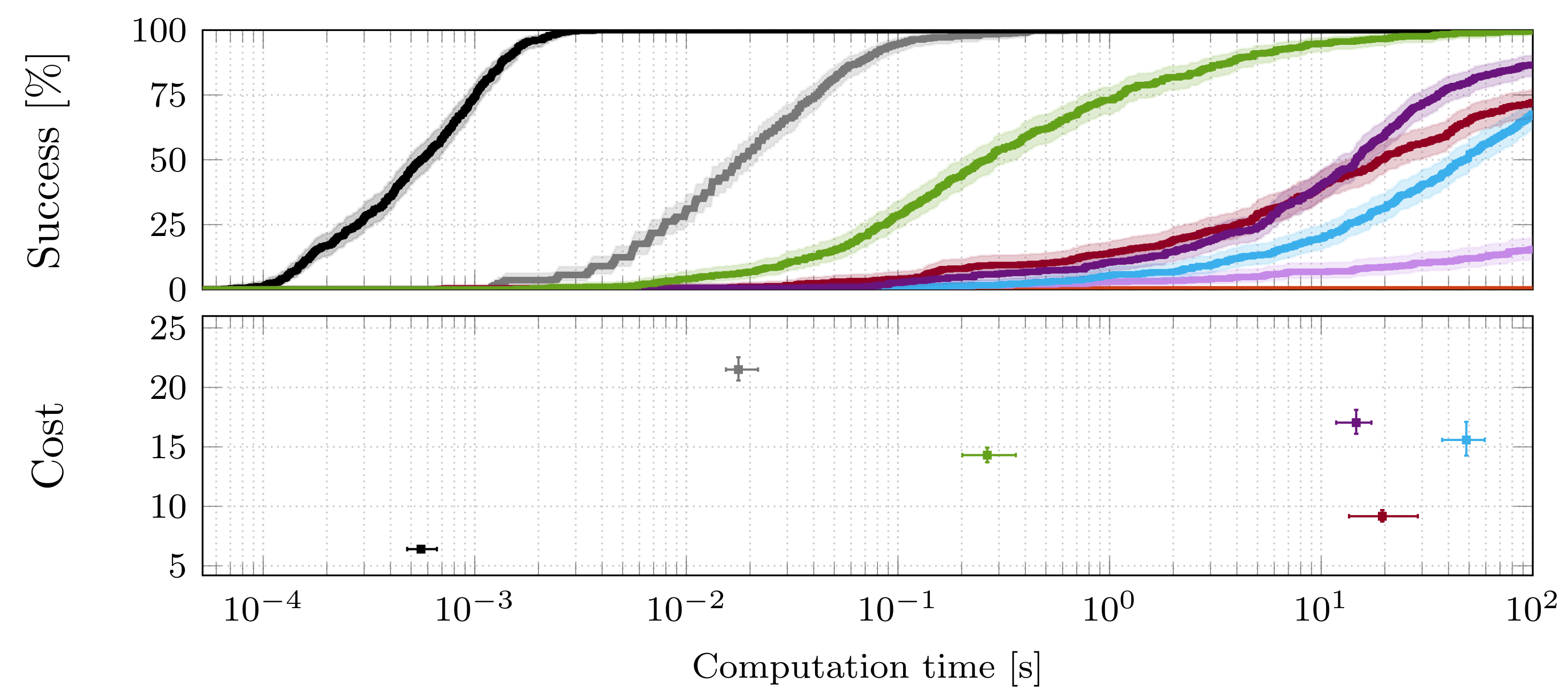}
    	\includegraphics[width=\linewidth]{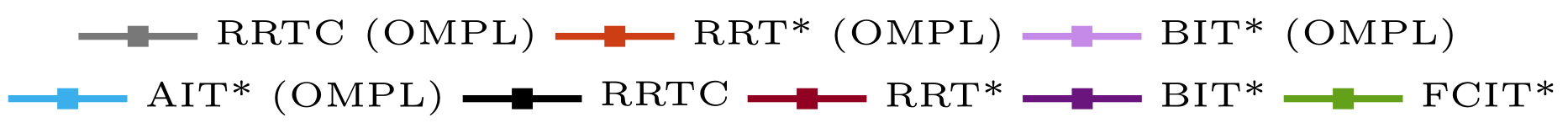}
        \caption{Planning Results}
    \end{subfigure}

	\caption{
        Results for the 7-DoF Panda~\cite{panda_approx} on the \emph{cage} environment from the MotionBenchMaker~\cite{mbm} dataset (\cref{sec:exp}).
        \textbf{(a)} Final computed paths for RRT-Connect (pink), RRT-Connect with standard path simplification heuristics applied (orange), and FCIT* (green) on the \emph{cage} problem evaluated in \cref{fig:converge:a}.
        \textbf{(b)} Planning results of all planners on all problems in the \emph{cage} environment.
		The upper plot shows the percentage of runs that found a solution at a given time with Clopper-Pearson 99\% confidence intervals.
		The lower plot shows the median initial path length with nonparametric 99\% confidence intervals.
        The only tested planner that finds initial solutions faster than \acs{FCIT*} in this environment is RRT-Connect, which is not an \acs{ASAO} algorithm and cannot improve its initial solution with additional computational time.} \label{fig:cage:all}
\end{figure}

If connectivity in a planner's sampled approximation is too low, then the graph maintained by the planner \textit{almost never} (i.e., with probability zero) contains a solution~\cite{prmstar,critical,rgg}, but if it is too high, the graph becomes expensive to search due to the high branching factor and resulting high number of edges.
As such, although incremental sampling avoids the need for \textit{a priori} approximations, sampling-based \acs{ASAO} planners still require a user-defined connectivity metric between samples (e.g., connection radius) for efficiency.

A significant body of work has refined the bounds necessary for almost-sure connectivity~\cite{fmt,detao,critical,enet}, but using these bounds has practical tradeoffs.
Considering only a subset of possible edges requires fewer edge evaluations but makes \emph{incomplete} use of the set of samples, potentially lowering the quality of the best solution that can be found without additional sampling.
Implementing a partially connected \acs{RGG} also requires nearest-neighbours structures to more efficiently query the incident edges of a given sample.
These edge evaluations and nearest-neighbour queries traditionally dominate execution time in sampling-based motion planning~\cite{bottleneck,lazyprm}.
\squeezeWords

\ac{VAMP}~\cite{vamp} has reduced the computational cost of edge evaluations using \ac{SIMD} parallelism, drastically accelerating solution times for feasible (i.e., satisficing) motion planning~\footnote{
	 The tracing compiler used in VAMP to generate interleaved collision checking code is applicable to any system with analytic forward kinematics, although the performance gain may vary based upon the application or system in question.
	 \squeezeWords
}, with RRT-Connect performing up to 500x faster than the previous state of the art~\cite{ompl,moveit}. %

The original \acs{VAMP} work did not address \acs{ASAO} motion planning.
This paper shows that its radically decreased edge evaluation cost can also guide algorithmic advances on this class of problems.
Specifically, we revisit the assumptions that edge evaluations are computationally expensive, and consequently that fully connected graphs are prohibitively expensive to search because of their high branching factor and the resulting large number of edges to evaluate.

This insight leads us to \ac{FCIT*}, an \ac{ASAO} planning algorithm that accelerates motion planning by searching a fully connected graph.
It does this efficiently by leveraging \ac{SIMD} parallelism to reduce the cost of edge evaluations and using a distributed edge queue to address the high branching factor of fully connected graphs.
These inexpensive edge evaluations allow \ac{FCIT*} to build an approximation with maximum connectivity (i.e., a fully connected, or complete, graph) instead of limiting connections to near a theoretical lower bound required for probabilistic guarantees.
This removes the need for costly nearest-neighbours structures and improves convergence.
The benefits of \ac{FCIT*} are demonstrated on the \ac{MBM}~\cite{mbm} dataset, where it finds better solutions faster than all tested non-\acs{VAMP} \ac{ASAO} algorithms (e.g.,~\cref{fig:cage:all}) and solves more problems, with better solutions, than all tested \ac{VAMP} \ac{ASAO} algorithms (\cref{plan_results}).
\squeezeWords

\section{Related Work} \label{sec:lit}

Informed graph-based searches, such as A*~\cite{astar}, search a discrete graph approximation of a continuously valued search space in order of potential solution quality using an admissible  heuristic (i.e., an underestimate of the true cost).
A* is both \emph{resolution optimal}, finding the optimal solution in an approximation if one exists, and \emph{optimally efficient}, expanding the minimum number of vertices to find the resolution optimal solution, but requires the continuous search space be discretized \textit{a priori}.

Sampling-based planners, like \ac{PRM}~\cite{prm} and \ac{RRT}~\cite{rrt},
place samples to create an approximation of the search space.
\ac{PRM} first builds a graph before searching it for a solution to a specific pair of vertices, while \ac{RRT} incrementally constructs a tree from the starting vertex until it is connected to the goal, and then extracts the solution.
These planners are \emph{anytime} in their approximation by sampling incrementally, avoiding the need for \textit{a priori} approximation.

Bidirectional planners like RRT-Connect~\cite{rrtc} perform simultaneous searches from the start and goal by building two trees to explore the search space and trying to connect them.
This bidirectional search performs well on many difficult problems and results in fast initial solution times, but is not \ac{ASAO} and provides no guarantees on solution quality.

Lazy planners, like LazySP~\cite{lazysp}, Lazy-PRM~\cite{lazyprm}, and Lower Bound Tree-RRT~\cite{lbtrrt}, reduce planning time by delaying costly operations like edge evaluation until necessary.
Edges are only evaluated when they are a part of a potential solution or to satisfy optimality bounds, reducing the number of these costly operations but requiring that the search be restarted after finding an invalid edge along the path.

Anytime \ac{ASAO} planning allows for an initial solution to be found quickly on a sparse approximation of the search space, and then higher quality solutions to be found as the resolution of the approximation increases with additional sampling.

RRT*~\cite{prmstar}, BIT*~\cite{bit}, and other similar \ac{ASAO} planners are anytime in their approximation, doing away with the need for \textit{a priori} approximation and iteratively sampling the search space and searching the graph.
BIT* is also informed in its search, using a heuristic to evaluate edges in order of potential solution quality.
This avoids unnecessary computational costs and improves planning convergence, allowing \ac{BIT*} to efficiently search its increasingly accurate \ac{RGG} approximation.

Planners like \ac{ABIT*}~\cite{abit}, \ac{GBIT*}~\cite{gbit}, and \ac{G-RRT*}~\cite{grrt} leverage a greedy heuristic search to better exploit the current approximation and find an initial solution faster than \ac{BIT*}. 
This avoids the high cost of finding an optimal initial solution by computing one that is `good enough' and then later searching the approximation for the resolution optimal solution.
This greedy search helps balance exploitation and exploration, but is done to avoid computationally expensive edge evaluations.

Lazy bidirectional \ac{ASAO} planners, like AIT*~\cite{ait}, use a reverse search to calculate a heuristic based on the current approximation that is then used to order the forward search.
This extra effort to calculate an accurate heuristic helps reduce the number of computationally expensive edge evaluations, improving the time to find initial solutions.

Edge evaluations can also be reduced by considering fewer outgoing edges per vertex.
Significant research has focused on deriving the lower bound of connectivity necessary to maintain almost-sure asymptotic optimality~\cite{fmt,detao,critical,enet}.
This reduces computational costs by considering fewer edges but will not fully exploit the current approximation and can lead to lower-quality solutions for a given set of samples.
If this connectivity is too low then the planner will quickly exploit the approximation but \textit{almost never} find a valid path\cite{prmstar,critical, rgg}, and if it is too high then the graph will almost surely contain a high quality solution but becomes increasingly expensive to search as the number of edges increases.

Searching an approximation with high connectivity (i.e. high branching factor) considers many suboptimal edges.
\ac{PEA*}~\cite{pea} generates all outgoing edges from a vertex, discarding those that are not promising, while \ac{EPEA*}~\cite{epea} uses \textit{a priori} information to generate only promising edges.
These graph-based searches may expand a vertex several times when searching a given approximation.
\squeezeWords

\begin{algorithm}[tb]
	\caption{FCIT*}\label{algo1}
	\SetInd{0.2em}{0.23em}
	$X_{\text{smpl}} \gets \{x_g\}$;
    $T \equiv \{V,E\}$;
	$V \gets \{x_\text{start}\}$\;
	
	$E \gets \emptyset$;
	$E_\text{invalid} \gets \emptyset$;
	$Q_\text{open} \gets \emptyset$; \label{algo1:oq1}
	$Q_\text{local}[x_\text{start}] \gets \emptyset$\; \label{algo1:sq1}
	
	\While{$\textbf{not } \text{Done}$}{
		$Q_\text{local}[x_\text{start}] \gets \{ (x_\text{start}, x \in X_\text{smpl}) \; | \; x \neq x_\text{start} \}$\; \label{algo1:sq2}
		$Q_\text{open} \gets \{\text{NextBestEdge}(x_\text{start})\}$\; \label{algo1:oq2}
		
		\While{$Q_\text{open} \not\equiv \emptyset$}{
			$(x_p,x_c) \gets \underset{(x,y)\in Q_\text{open}}{\text{argmin}} (\hat{f}(x,y))$\; \label{algo1:popedge}
			$Q_\text{open} \stackrel{-}\gets \{x_p,x_c\}$\;
			$Q_\text{open} \stackrel{+}\gets \{\text{NextBestEdge}(x_p)\}$\;
			
			\uIf{$p(x_c) \equiv x_p$}{
				$Q_\text{local}[x_c] \gets \{ (x_c, x \in X_\text{smpl}) \; | \; x \neq x_c \}$\; \label{algo1:sq3}
				$Q_\text{open} \stackrel{+}\gets \{\text{NextBestEdge}(x_c)\}$\;
			}
   
			\ElseIf{$\hat{f}(x_p,x_c) \leq \underset{x_g \in X_g}{\min} \{g_T(x_g)\}$}{
				\If{$g_T(x_p)+\hat{c}(x_p,x_c) \leq g_T(x_c)$}{
                    \If{$\text{IsValid}(x_p,x_c)$}{
                        \If{$g_T(x_p)+c(x_p,x_c)+\hat{h}(x_c) \leq \underset{x_g \in X_g}{\min} \{g_T(x_g)\}$} {
                            
                            \If{$g_T(x_p)+c(x_p,x_c) \leq g_T(x_c)$}{
                                \If {$x_c \in V$}{
                                  $E \stackrel{-}\gets (p(x_c), x_c)$\;
                                }\Else{
                                  $V \stackrel{+}\gets x_c$\;
                                  $Q_\text{local}[x_c] \gets \{ (x_c, x \in X_\text{smpl}) \; | \; x \neq x_c \}$\; \label{algo1:distributeedges}
                                }
                                
                                $E \stackrel{+}\gets (x_p, x_c)$\;
                                $Q_\text{open} \stackrel{+}\gets \{\text{NextBestEdge}(x_c)\}$\; \label{algo1:getbest}
                            }
                        }
                    }
                    \ElseIf{$(x_p, x_c) \not\in E_\text{invalid}$}{
                        $E_\text{invalid} \stackrel{+}\gets \{(x_p,x_c), (x_c,x_p)\}$\;
                    }
				}
			}
			\Else{
				$Q_\text{open} \gets \emptyset$\;
			}
		}
		$X_{\text{smpl}} \stackrel{+}\gets \text{AddSamples}(n)$\;
	}
	\Return{$T$}
 
\end{algorithm}

\begin{algorithm} [tb] 
	\caption{NextBestEdge($x_p$)}\label{algo2}
    \SetInd{0.2em}{0.23em}
	\While{$Q_\text{local}[x_p] \not\equiv \emptyset$}{
		$x_c \gets \underset{(x,y)\in Q_\text{local}[x_p]}{\text{argmin}} (\hat{f}(x,y))$\;
		$Q_\text{local}[x_p] \stackrel{-}\gets \{x_c\}$\;
		\If{$g_T(x_p) + \hat{c}(x_p,x_c) < g_T(x_c)$}{
			\Return{$(x_p, x_c)$}\;
		}
	}
\end{algorithm}

In comparison, \ac{FCIT*} reduces the cost of edge evaluations with \ac{SIMD} parallelism and avoids nearest-neighbours structures by considering a fully connected graph.
This completely exploits the current set of samples and is done efficiently via an informed search over a distributed edge queue.
Unlike other VAMP algorithms, \ac{FCIT*} almost-surely converges asymptotically to the optimal solution with additional computational time. 
\ac{FCIT*}  orders its search by potential solution quality as in BIT*, but does not connect samples within a critical connection radius or require nearest-neighbours structures, instead considering every possible edge in the graph.
Unlike \ac{PEA*} and \ac{EPEA*}, \ac{FCIT*} generates and stores all outgoing edges for a vertex only once for a given approximation.

\section{\acf{FCIT*}} %

\ac{FCIT*} extends the \ac{BIT*} algorithm by leveraging the hardware-accelerated parallelized approach to edge validation of \ac{VAMP}~\cite{vamp} and revisiting the assumption that
a high branching factor and the high cost of edge evaluation makes fully connected graphs prohibitively expensive to construct and search.
It builds and searches a fully connected approximation of the continuous search space, as opposed to relying on a sparsely connected approximation built using nearest-neighbours structures as in \ac{BIT*}.
Building a fully connected approximation makes complete use of all placed samples, potentially containing a higher quality solution than could be found in an approximation with connectivity near a theoretical lower bound~\cite{critical}.
\squeezeWords

Constructing and searching a fully connected graph also removes the need to maintain expensive nearest-neighbours structures by instead considering all possible edges between sampled states.
These edges \emph{may} all be evaluated in the limit, but in practice many will not be because they exist in disconnected components, or belong to a more expensive path than the current best solution.

\ac{FCIT*}'s search is ordered by potential solution quality (\cref{algo1} \cref{algo1:popedge}) as in \ac{BIT*}.
Unlike \ac{BIT*}, it distributes its ordered queue of unprocessed (i.e., open) edges across a set of local queues stored by each vertex (\cref{algo1} \cref{algo1:distributeedges}).
Only the most promising edges from each local queue are added to the global queue and sorted (\cref{algo1} \cref{algo1:getbest}), resulting in a more efficient search.
Pseudocode for \ac{FCIT*} is provided in \cref{algo1,algo2}.
\squeezeWords

\begin{figure*}[t!]
	\subfloat[table pick (OMPL)]{%
		\includegraphics[width=.49\linewidth]{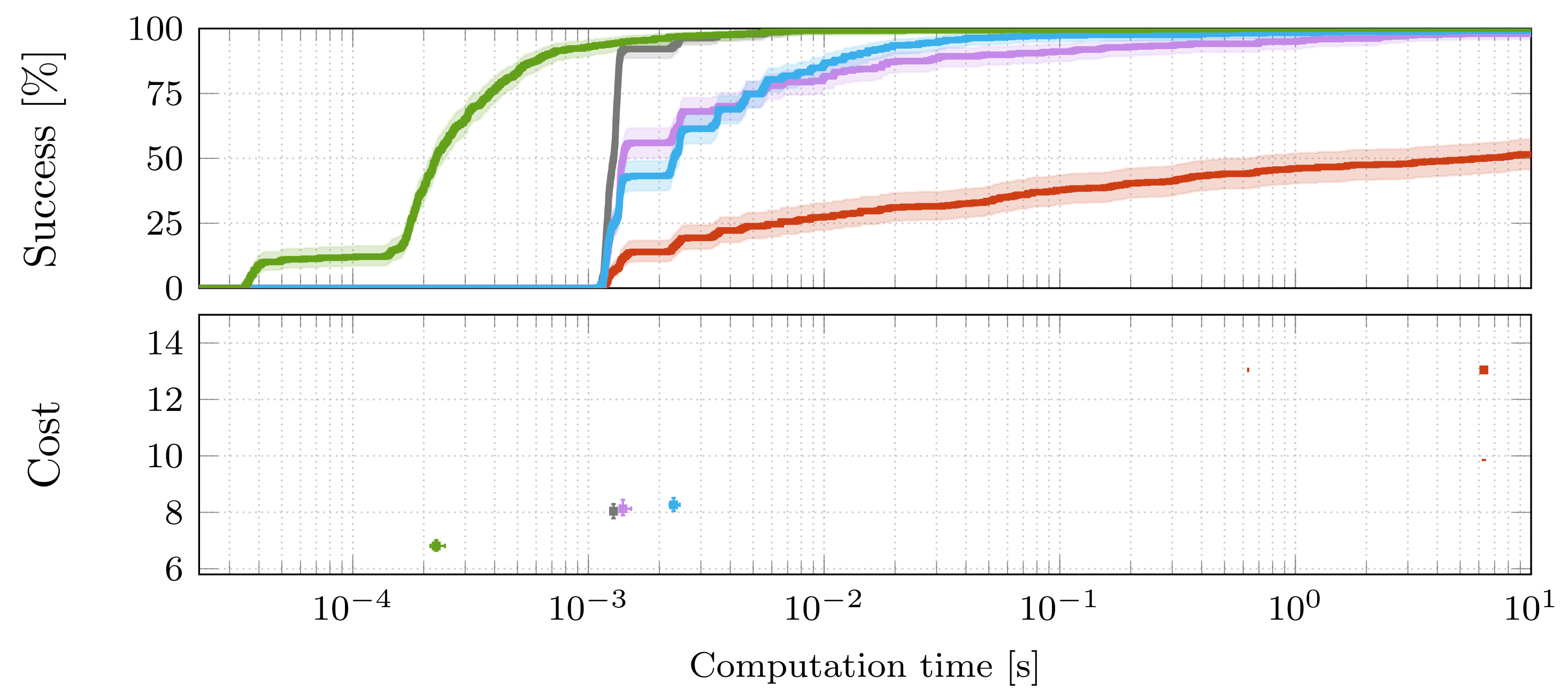}%
	}\hfill
	\subfloat[table pick (VAMP)]{%
		\includegraphics[width=.49\linewidth]{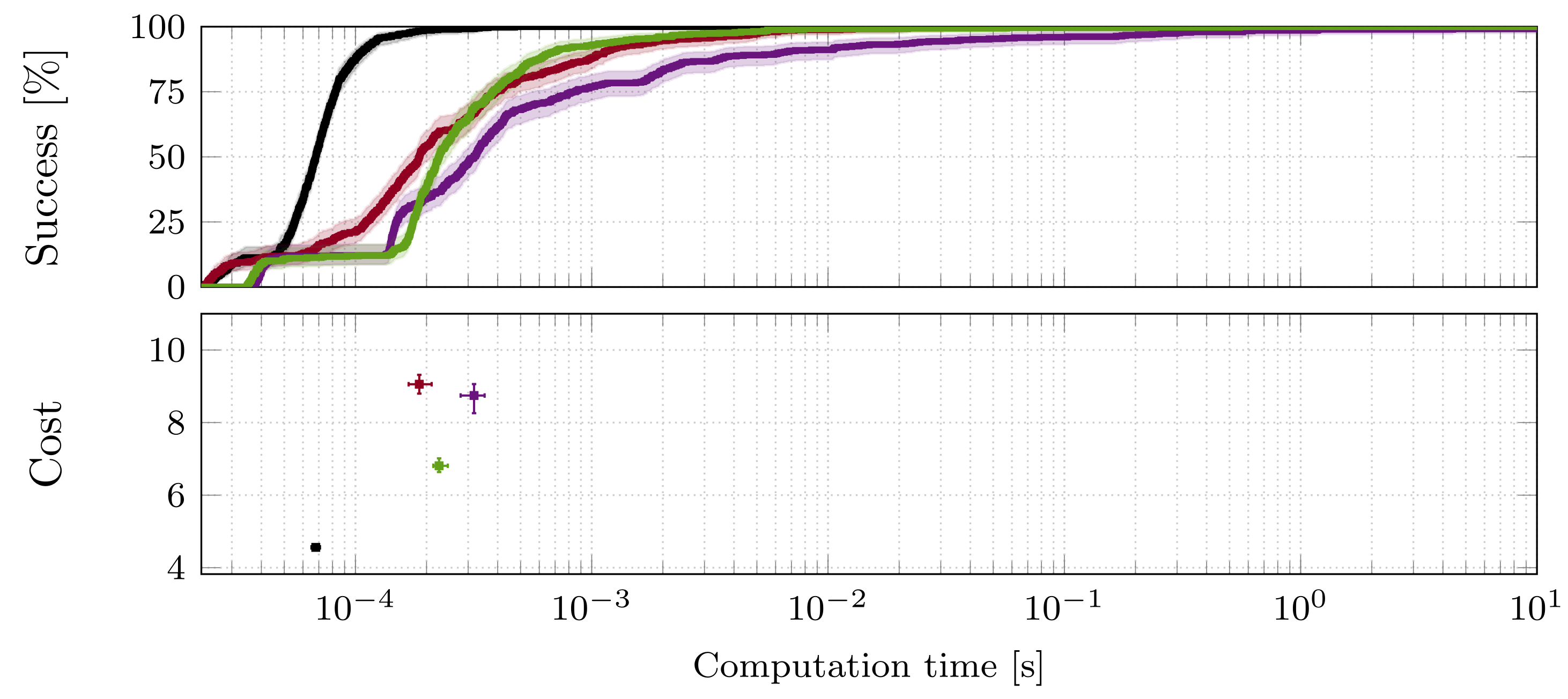}%
	}\\
	\subfloat[bookshelf small (OMPL)]{%
		\includegraphics[width=.49\linewidth]{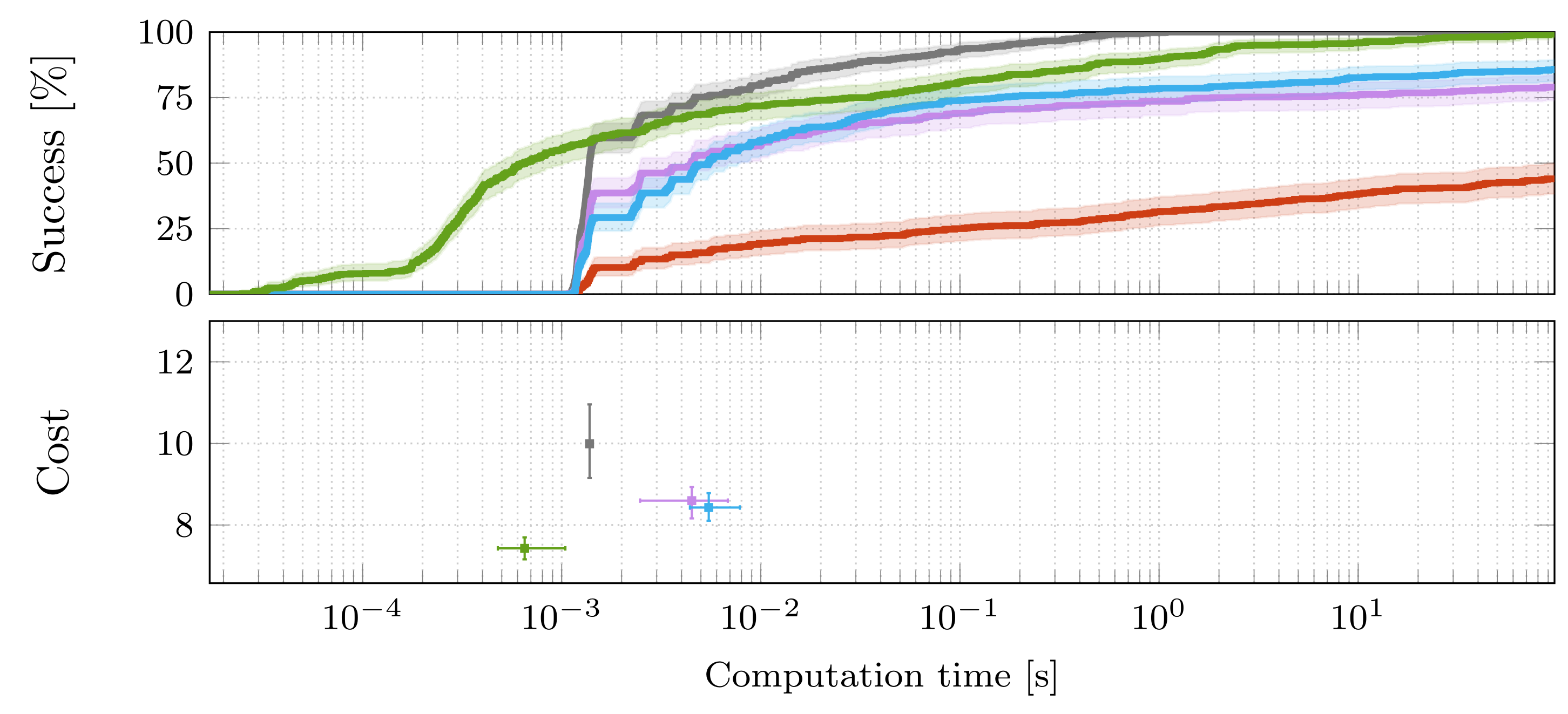}%
	}\hfill
	\subfloat[bookshelf small (VAMP)]{%
		\includegraphics[width=.49\linewidth]{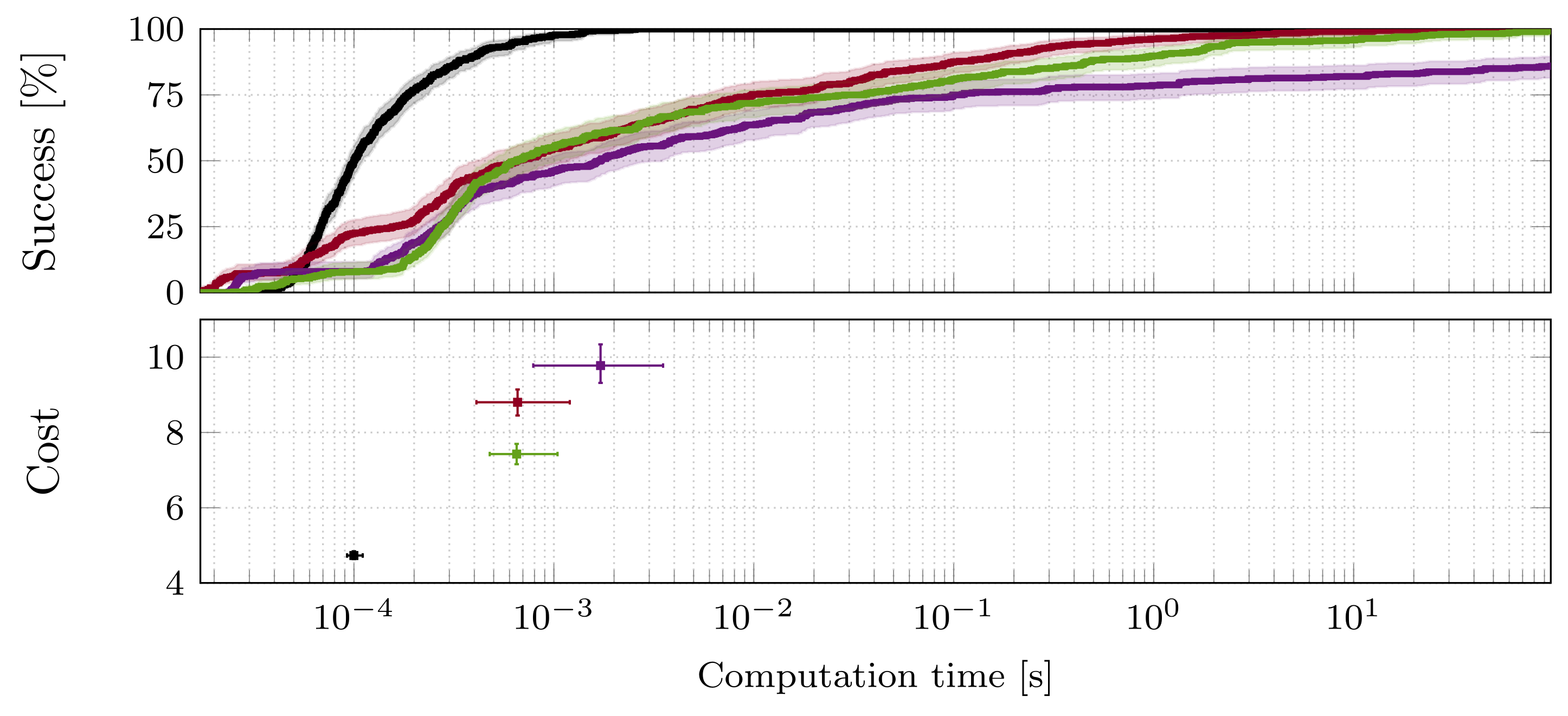}%
	}\hfill
	\subfloat{%
		\includegraphics[width=\linewidth]{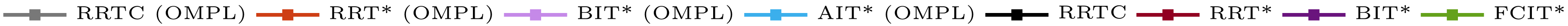}%
	}
	\caption{ Results for the 7-DoF Panda~\cite{panda_approx} on all problems in the \emph{table pick} and \emph{bookshelf small} environments (\cref{sec:exp}).
		Plots \textbf{(a)} and \textbf{(c)} compare \ac{FCIT*} to several OMPL planners, while plots \textbf{(b)} and \textbf{(d)} compare \ac{FCIT*} to several VAMP planners.
		For each plot, the top shows the percentage of runs that found a solution at any given time with Clopper-Pearson 99\% confidence intervals,
		and the bottom shows the median initial path length with nonparametric 99\% confidence intervals.
        The only tested planner that finds initial solutions significantly faster than \acs{FCIT*} in these environments is RRT-Connect, which is not an \acs{ASAO} algorithm and cannot improve its initial solution with additional computational time.}
	\label{fig:2:all}
\end{figure*}

\subsection{Notation and Preliminaries}

We denote the search space by $X \subseteq \mathbb{R}^n$ and the invalid and valid subsets as $X_\text{invalid} \subseteq X$ and $X_\text{valid} \coloneqq \mathtt{closure}(X \setminus X_\text{invalid})$, respectively.
Let $x \in X$ be a state, $x_\text{start} \in X_\text{valid}$ be the start state, and $x_\text{goal} \in X_\text{valid}$ be the goal state.
The set of all sampled valid states is denoted as $X_{\text{smpl}} \subset X_\text{valid}$.
We store the search as a tree, $T = (V,E)$, comprising a set of vertices from the sampled states, $V \subseteq X_{\text{smpl}}$, and a set of edges, $E \subseteq V \times V$.
Each edge connects two states, $x_p, x_c \in X_{\text{smpl}}$, which we refer to as the edge's parent and child, respectively.
\squeezeWords

The planner also tracks the set of invalidated edges, denoted by $E_\text{invalid} \subseteq V \times X_{\text{smpl}}$, and maintains a queue of open edges denoted as $Q_\text{open} \subseteq V \times X_{\text{smpl}}$.
Each vertex, $x \in V$, has an associated local outgoing edge queue stored in a lookup table, $Q_\text{local}(x)$, such that $Q_\text{local}(x) \coloneqq \{ (x,y \in X_\text{smpl}) \; | \; y \neq x \}$.
The functions $p(x)$ and $g_T(x)$ respectively calculate the parent and cost-to-come from the start through the tree, $T$, for a state $x \in X_{\text{smpl}}$.
These functions return infinity if the state is not in the tree, i.e., $x \not\in V$.

Following the formulation in \ac{BIT*}~\cite{bit}, the function $c: X \times X \to [0, \infty)$ represents the computed edge cost between two states.
The function $\hat{c}: X \times X \to [0, \infty)$ is an admissible estimate of this edge cost, where $\forall x, y \in X, \hat{c}(x, y) \leq c(x, y)$.
The function $\hat{h}: X \to [0, \infty)$ represents the estimated cost-to-go from the state $x$ to the goal, e.g., $\hat{h}(x) = \min_{x_g \in X_g} (\hat{c}(x, x_g))$.
The function $\hat{f}: V \times X \to [0, \infty)$ represents an admissible heuristic estimate of the cost of a solution constrained to pass through an edge given the current tree.
It is calculated as the sum of the current cost-to-come through the tree, the estimated edge cost, and the estimated cost-to-go, i.e., $\hat{f}(x_p,x_c) \coloneqq g_T(x_p)+\hat{c}(x_p, x_c)+\hat{h}(x_c)$.

Let $A$ and $B$ be two sets.
The notation $A \stackrel{+}\gets B$ is shorthand for the operation $A \gets A \cup B$, and $A \stackrel{-}\gets B$ is shorthand for the operation $A \gets A \setminus B$.
The number of states sampled in each batch is denoted by $n$.

\subsection{Local Edge Queue} \label{sec:local}

\ac{FCIT*} maintains a sorted open queue of edges to be expanded, $Q_\text{open}$, ordered by potential solution cost.
This open queue has to be sorted each time a new edge is added to it.
As more edges are added, it becomes longer and takes more time to sort.
To reduce the computational cost of this frequent sort, \ac{FCIT*} distributes the total set of open edges such that each vertex, $x \in V$, keeps its own local queue of outgoing edges, $Q_\text{local}(x)$.
These local edge queues are sorted \emph{once} each by the admissible heuristic estimate of solution cost, $\hat{f}$, through each edge.
The open queue, $Q_\text{open}$, is then populated with the most promising edge from each vertex's local edge queue (\cref{algo1}~\cref{algo1:oq2}).
The open queue is thus the ordered set of only the \emph{best} open edge outgoing from each vertex, $Q_\text{open} \coloneqq \{ (x, y) \in V \times X_\text{smpl} \mid (x,y) \leq \argmin_{(x,y)\in Q_\text{local}(x)} \{\hat{f}(x,y)\}\}$.
When an edge outgoing from a vertex is removed from the open queue, it is replaced by the next best outgoing edge from that vertex's local queue that could potentially improve the current tree (\cref{algo1} \cref{algo1:getbest}, \cref{algo2}).
This ensures that the open queue always contains the most promising unevaluated outgoing edges from all the vertices in the search tree.
\squeezeWords

Each vertex in a fully connected RGG has a number of potential outgoing edges equal to $|X_{\text{smpl}}|-1$, where $|\cdot|$ is the cardinality of a set.
Expanding vertices in the tree quickly inflates the total number of open edges, in the worst case increasing it to $|X_{\text{smpl}}|^2$ elements.
These local queues reduce the worst case size of the open queue from $|X_{\text{smpl}}|^2$ to $|X_{\text{smpl}}|$.
Expanding a new vertex only adds one new edge to the open queue to be frequently sorted, storing the other $|X_{\text{smpl}}|-2$ edges in that vertex's local queue and sorting them only once.
The frequent cost of sorting the open queue, $Q_\text{open}$, is thus reduced, and the one-time cost of sorting a given vertex's local queue, $Q_\text{local}(x)$, is amortized over the runtime of the~planner.

\subsubsection{Nearest-Neighbours Structures}
The set of outgoing edges from a given vertex is determined by the connectivity of the graph, i.e., the neighbouring vertices with which it shares edges.
In contrast to traditional algorithms, which typically maintain an expensive nearest-neighbours structure, finding neighbours is trivial in a fully connected graph since the neighbours of any given vertex are every other sampled state in the graph.
\ac{FCIT*} therefore iterates over all sampled points, $X_{\text{smpl}}$, when populating a vertex's local edge queue, $Q_\text{local}(x) \coloneqq \{ (x,y \in X_\text{smpl}) \; | \; y \neq x \}$, avoiding the computational cost of maintaining a nearest-neighbours structure and reducing planning time.

\begin{figure}
	\centering
	\begin{subfigure}[b]{\linewidth}
		\includegraphics[width=0.99\linewidth]{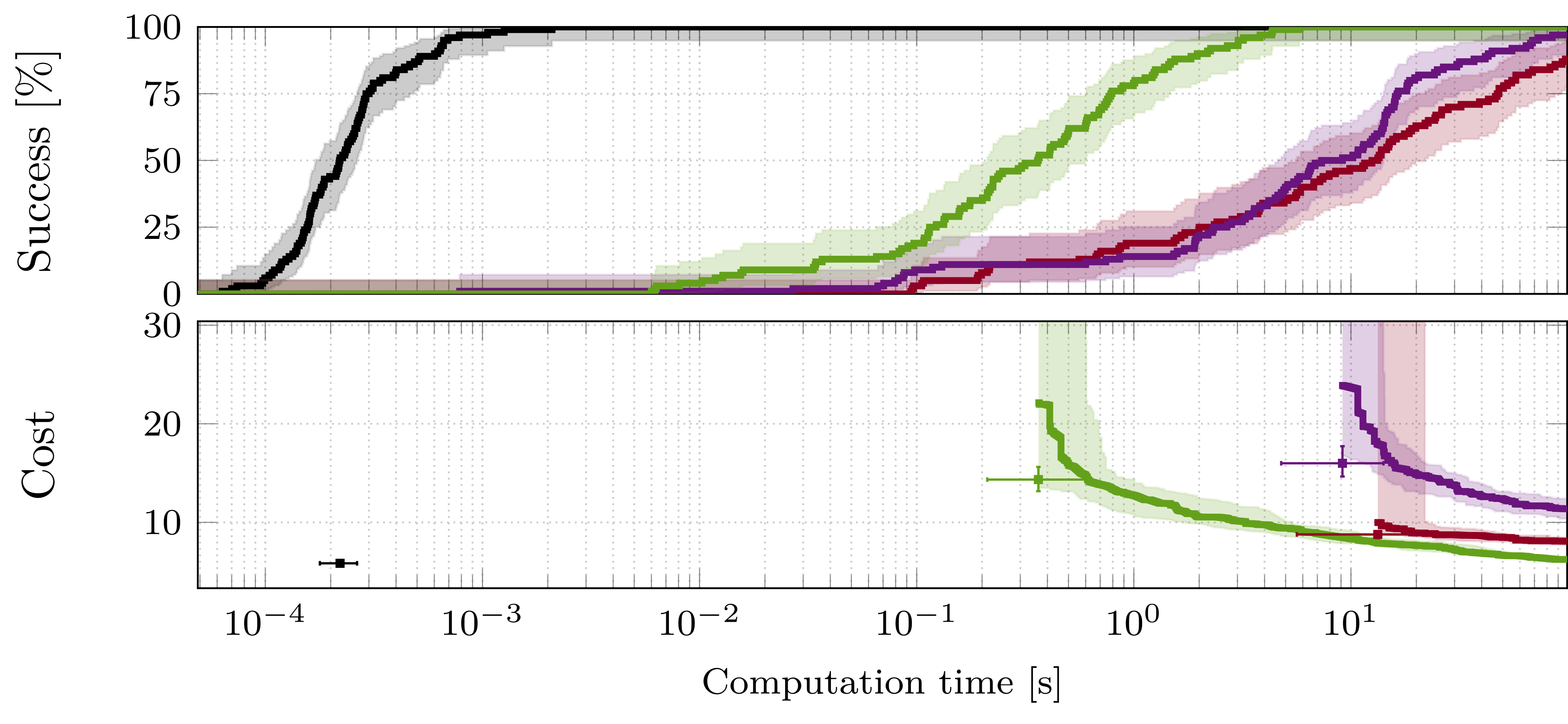}
		\caption{cage}
        \label{fig:converge:a}
	\end{subfigure}
	
	\begin{subfigure}[b]{\linewidth}
		\includegraphics[width=0.99\linewidth]{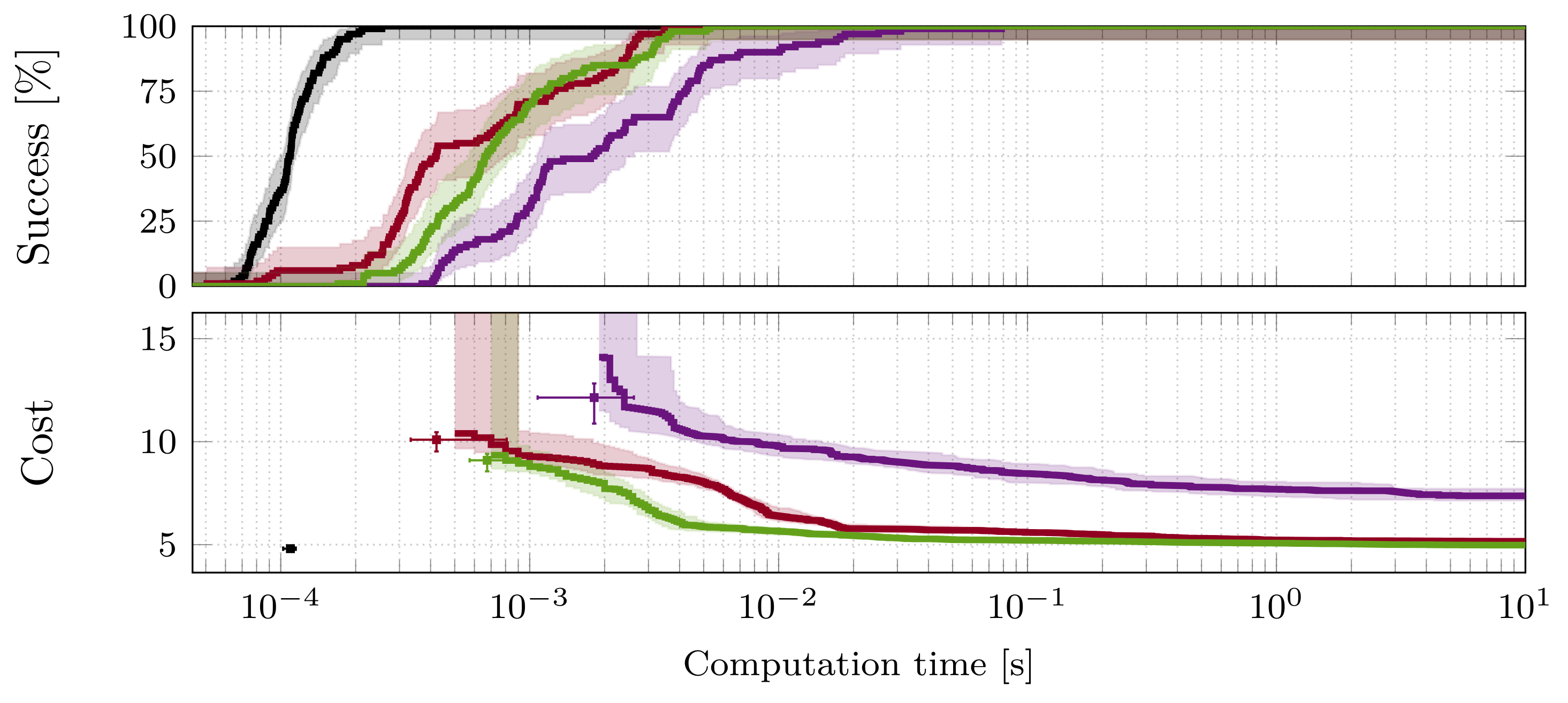}
		\caption{bookshelf small}
        \label{fig:converge:b}
	\end{subfigure}
	
	\begin{subfigure}[b]{0.75\linewidth}
		\includegraphics[width=1\linewidth]{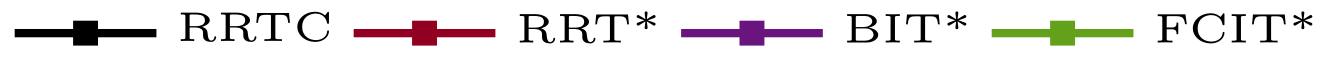}
	\end{subfigure}
	
	\caption{
		Convergence results for VAMP planners with the 7-DoF Panda~\cite{panda_approx} over 100 trials on a single problem from the \emph{cage} \textbf{(a)} and \emph{bookshelf small} \textbf{(b)} environments from the MotionBenchMaker~\cite{mbm} dataset (\cref{sec:exp}).
		For each plot, the top shows the percentage of runs that found a solution by a given time with Clopper-Pearson 99\% confidence intervals,
		and the bottom shows the median initial path length and median path length over time with nonparametric 99\% confidence intervals.
        The only tested planner that finds initial solutions significantly faster than \acs{FCIT*} in these environments is RRT-Connect, which is not \acs{ASAO} and cannot improve its initial solution with additional computational time.
	} \label{fig:converge}
\end{figure}

\subsection{Analysis}

This paper uses Definition 24 in \cite{prmstar} as the definition of almost-sure asymptotic optimality.
Note that any sampling-based planner is almost-surely asymptotically optimal if (i) its underlying graph almost-surely contains an asymptotically optimal path, and (ii) its underlying graph-search is resolution optimal.
These conditions are sufficient but not necessary.

Since there is a non-zero probability of sampling every state in the search space, the probability that the solution found by \ac{FCIT*} will asymptotically converge to the optimum approaches one as the number of samples approaches infinity, regardless of whether the sampling used is pseudorandom or deterministic~\cite{detao}.
This holds true for a fully connected graph since it trivially satisfies the lower bound on connectivity, as demonstrated by the simplified-\ac{PRM}~\cite{prmanalysis} \ac{ASAO} proof with an infinite r-disc graph~\cite{prmstar}.
Since the search is performed in an informed order as in \ac{BIT*}, it is also \emph{resolution optimal}~\cite{bit}, finding the best possible solution with respect to the current approximation.
\ac{FCIT*} is therefore almost surely asymptotically optimal.

\subsection{Implementation}

In practice, each vertex locally stores its set of invalid edges, similar to its local edge queue, instead of maintaining a global list.
The cost-to-come function, $g_T(x)$, and parent function, $p(x)$, are implemented as lookups, storing and updating the values to reduce time per iteration.
The open and local queues are both implemented as sorted structures.

\section{Experiments} \label{sec:exp}

\ac{FCIT*} was evaluated against \ac{OMPL}~\cite{ompl} and VAMP~\cite{vamp} baselines.
We compared to \ac{OMPL} implementations of RRT-Connect, RRT*, BIT*, and AIT*, as well as VAMP implementations of RRT-Connect, RRT*, and BIT*.
Note that the VAMP implementation of BIT* uses the same local edge queue presented in \cref{sec:local}, but performs a traditional $r$-disc nearest-neighbours search rather than using a fully connected graph.
The VAMP implementation of RRT-Connect is a dynamic domain~\cite{dynamic} balanced~\cite{balanced} RRT-Connect~\cite{capt}.
The reported initial solution costs and times for RRT-Connect include path smoothing with randomized shortcutting~\cite{shortcut1, shortcut2} and B-spline smoothing~\cite{bspline}.

The planners were tested on the \ac{MBM}~\cite{mbm} dataset, which consists of 7 difficult planning environments each containing 100 pregenerated problems.
These environments cover a range of planning problems, including reaching (\textit{bookshelf tall}, \textit{bookshelf small}, and \textit{bookshelf thin}), highly constrained reaching (\textit{box} and \textit{cage}), and tabletop manipulation (\textit{table pick} and \textit{table under pick})\footnote{One of the problems in \textit{table pick} is invalid with respect to the robot simulation and is disregarded, giving these experiments 699 total problems.}.

\rowcolors{1}{white}{lightergray}
\begin{table*} [tb] 
    \caption{
		Summary of all planning results.
		The \textbf{top} results in each row are for the VAMP implementation of the given planner, while the \textbf{bottom} results are for the \ac{OMPL} implementation of that planner.
      The only tested planner that solves more problems than \acs{FCIT*} is RRT-Connect, which is not an \acs{ASAO} algorithm and cannot improve its initial solution with additional computational time. Only VAMP RRT-Connect reliably finds initial solutions faster than \ac{FCIT*}.
		Each result indicates the percentage of problems solved in the given environment (bold), the median initial solution time across all problems on that environment, and the median initial path length across all problems on that environment.
	}\label{plan_results}
	\begin{tabular}{llllllll} 
		\toprule
		\textbf{Planner} & 
		\textbf{bookshelf small} &
		\textbf{bookshelf tall} & 
		\textbf{bookshelf thin} & 
		\textbf{table pick} & 
		\textbf{table under pick} & 
		\textbf{box} & 
		\textbf{cage} \\
		\midrule
		RRTConnect & 
		\tablecell{\textbf{100\%} 0.1ms (4.7)}{\textbf{100\%} 1.4ms (10.0)} & %
		\tablecell{\textbf{100\%} 0.1ms (4.7)}{\textbf{100\%} 1.3ms (8.8)} & %
		\tablecell{\textbf{100\%} 0.1ms (4.6)}{\textbf{100\%} 1.3ms (8.7)} & %
		\tablecell{\textbf{100\%} 0.1ms (4.6)}{\textbf{100\%} 1.3ms (8.0)} & %
		\tablecell{\textbf{100\%} 0.1ms (6.0)}{\textbf{100\%} 1.3ms (12.6)} & %
		\tablecell{\textbf{100\%} 0.2ms (4.4)}{\textbf{100\%} 1.4ms (11.3)} & %
		\tablecell{\textbf{100\%} 0.6ms (6.4)}{\textbf{100\%} 18ms (21.5)} \\ %
		
		RRT* & 
		\tablecell{\textbf{99\%} 0.9ms (8.8)}{\textbf{44\%} $\infty$ ($\infty$)} & %
		\tablecell{\textbf{100\%} 0.9ms (8.7)}{\textbf{39\%} $\infty$ ($\infty$)} & %
		\tablecell{\textbf{100\%} 0.7ms (9.3)}{\textbf{51\%} 8827ms (8.7)} & %
		\tablecell{\textbf{100\%} 0.2ms (9.1)}{\textbf{52\%} 6304ms (13.0)} & %
		\tablecell{\textbf{100\%} 0.3ms (14.9)}{\textbf{33\%} $\infty$ ($\infty$)} & %
		\tablecell{\textbf{100\%} 1.7ms (8.1)}{\textbf{17\%} $\infty$ ($\infty$)} & %
		\tablecell{\textbf{72\%} 19460ms (9.2)}{\textbf{0\%} $\infty$ ($\infty$)} \\ %
		
		BIT* & 
		\tablecell{\textbf{86\%} 1.7ms (9.8)}{\textbf{79\%} 4.5ms (8.6)} & %
		\tablecell{\textbf{95\%} 1.7ms (9.3)}{\textbf{91\%} 3.4ms (8.6)} & %
		\tablecell{\textbf{98\%} 1.8ms (9.4)}{\textbf{93\%} 4.4ms (8.3)} & %
		\tablecell{\textbf{99\%} 0.3ms (8.7)}{\textbf{98\%} 1.4ms (8.1)} & %
		\tablecell{\textbf{100\%} 0.4ms (12.4)}{\textbf{98\%} 2.2ms (11.4)} & %
		\tablecell{\textbf{100\%} 1.8ms (10.5)}{\textbf{98\%} 7.7ms (9.9)} & %
		\tablecell{\textbf{87\%} 14644ms (17.0)}{\textbf{16\%} $\infty$ ($\infty$)} \\ %
		
		AIT* & 
		\tablecell{---}{\textbf{86\%} 5.5ms (8.4)} & %
		\tablecell{---}{\textbf{95\%} 4.4ms (8.5)} & %
		\tablecell{---}{\textbf{98\%} 4.5ms (8.3)} & %
		\tablecell{---}{\textbf{99\%} 2.3ms (8.3)} & %
		\tablecell{---}{\textbf{100\%} 3.4ms (11.5)} & %
		\tablecell{---}{\textbf{100\%} 11.1ms (10.3)} & %
		\tablecell{---}{\textbf{68\%} 48541ms (15.6)} \\ %
		
		\ac{FCIT*} & 
		\tablecell{\textbf{99\%} 0.7ms (7.4)}{---} & %
		\tablecell{\textbf{100\%} 0.4ms (7.5)}{---} & %
		\tablecell{\textbf{100\%} 0.5ms (7.8)}{---} & %
		\tablecell{\textbf{100\%} 0.2ms (6.8)}{---} & %
		\tablecell{\textbf{100\%} 0.4ms (10.7)}{---} & %
		\tablecell{\textbf{100\%} 0.9ms (8.1)}{---} & %
		\tablecell{\textbf{100\%} 264ms (14.3)}{---} \\ %
		
		\bottomrule
	\end{tabular}
\end{table*}

All experiments were run using a simulated 7-\ac{DoF} Panda robotic arm~\cite{panda_approx}.
Each problem was evaluated 5 times by each planner to mitigate the effect of machine noise on the results\footnote{All tests were run in Ubuntu 22.04 on a Intel i7-9750H CPU with 32GB of RAM, and the planning algorithms are implemented in C++17.}.
All planners use the default VAMP and \ac{OMPL} samplers with different seeding for each trial.
Planners were all given the same time constraints to evaluate each problem in a given environment.
The time constraints per environment were 10 seconds on \emph{box}, \emph{table pick}, \emph{table under pick}, \emph{bookshelf thin}, and \emph{bookshelf tall}; and 100 seconds on \emph{bookshelf small} and \emph{cage}.
\squeezeWords

Experimental results for all problems in a given environment are shown in \cref{fig:cage:all,fig:2:all}, and are summarized in \cref{plan_results}.
The time axis for all figures is in logarithmic scale.
Results for each environment are presented separately because all the environments in the \ac{MBM} dataset pose significantly different planning problems from each other.
While \cref{plan_results} includes initial solution results for all planners across all environments, \cref{fig:cage:all,fig:2:all} show results for all problems in \emph{cage}, \emph{table pick}, and \emph{bookshelf small}.
These results were chosen because they are the most indicative of relative qualitative planner performance.
\cref{fig:converge} shows convergence results for 100 trials of each VAMP planner on a single problem from both the \emph{cage} and \emph{bookshelf small} environments.
This figure omits \ac{OMPL} planner convergence results because \ac{FCIT*} finds initial solutions significantly faster and of higher quality than \emph{all} ASAO \ac{OMPL} planners (\cref{fig:cage:all},~\ref{fig:2:all}a, and~\ref{fig:2:all}c).

\section{Discussion}

\ac{FCIT*} revisits fundamental assumptions about \ac{ASAO} planning in light of the computationally inexpensive local motion validation introduced by \ac{VAMP}~\cite{vamp}.
Planners traditionally seek to reduce the number of edges in their approximation by setting connectivity near a theoretical lower bound.
This requires maintaining a computationally expensive nearest-neighbours structure.
Moreover, this limits the connectivity of the resulting graph, preventing solutions from being found without additional sampling.
\ac{FCIT*} is able to avoid this lower bound because of the reduced cost of edge evaluation, instead searching a fully connected approximation.
This removes the need for maintaining a nearest-neighbours structure, instead considering all possible edges in the graph in an informed order.
It uses this fully connected \ac{RGG} approximation to find better solutions with fewer required samples.

\ac{FCIT*} outperforms all other tested ASAO planners, both VAMP and \ac{OMPL}, on the difficult \emph{cage} environment (\cref{fig:cage:all}).
The only tested planners that finds initial solutions faster than \ac{FCIT*} in this environment are the VAMP and \ac{OMPL} implementations of RRT-Connect, which do not converge towards an optimal solution given additional planning time.
\squeezeWords

\ac{FCIT*} demonstrates better performance than other tested VAMP ASAO planners. 
\ac{FCIT*} consistently outperforms the VAMP BIT* planner in all environments, finding better initial solutions in less time (\cref{fig:2:all}) by fully exploiting the samples.
VAMP RRT* shows similar initial solution times to \ac{FCIT*} on the simpler environments, but consistently yields lower quality solutions, and performs much worse than \ac{FCIT*} in the difficult \emph{cage} environment, only solving 72\% of problems (\cref{plan_results}).
\squeezeWords

\ac{FCIT*} solves more problems than any other tested planner barring the two implementations of RRT-Connect, only occasionally failing to solve one difficult problem in the \emph{bookshelf small} environment.
It also outperforms \ac{OMPL} RRT-Connect on all environments other than \emph{cage}, finding faster initial solutions with additional ASAO guarantees.
The initial solution time for \ac{FCIT*} is consistently within an order of magnitude of that of \ac{VAMP} RRT-Connect on all environments except for the difficult \emph{cage} environment (\cref{plan_results}).
This environment seems well suited for bidirectional planners, as also evidenced by the performance of AIT* relative to the other \ac{OMPL} \ac{ASAO} planners.

\ac{FCIT*} outperforms all tested \ac{OMPL} \ac{ASAO} planners on all environments, finding higher quality solutions in less time.
All tested \ac{OMPL} planners show a delay before finding initial solutions (\cref{fig:2:all}).
This can be attributed to overhead in \ac{OMPL}, even though only planning time is reported for these planners.

We believe that the introduction of VAMP poses an open question of how to best leverage trivialized edge evaluation, since so much existing planning research has focused on avoiding these operations on the assumption that they are computationally expensive.
\ac{FCIT*} is an initial answer to this question, but it is clear there is further research to be done on the topic.
We are particularly interested in finding ways to apply the benefits of bidirectional search to \ac{FCIT*}, potentially closing the gap to RRT-Connect's performance on difficult environments.
\squeezeWords

\section{Conclusions}

Motion planning is an ongoing and important area of research in robotics.
This paper leverages VAMP to reduce the cost of edge evaluation and presents \ac{FCIT*}, the first fully connected, informed, anytime almost-surely asymptotically optimal planner.
\ac{FCIT*} leverages inexpensive edge evaluations to build and search a fully connected graph instead of limiting connections to near a theoretical lower bound.
This allows it to fully exploit all samples in a given approximation without requiring nearest-neighbours structures, instead considering every possible edge in the approximation and yielding better solutions in less time and with fewer samples placed.
\squeezeWords

The benefits of leveraging VAMP to search a fully connected graph are demonstrated on hundreds of problems across seven different planning environments.
\ac{FCIT*} demonstrates performance comparable to that of the fastest planner, VAMP's RRT-Connect, on almost all environments tested and with additional guarantees.
It outperforms all other tested VAMP and \ac{OMPL} ASAO planners, finding initial solutions faster, of higher quality, and more consistently, and outperforms \ac{OMPL}'s RRT-Connect on all but the most difficult class of problems tested, all while maintaining ASAO guarantees.

Information on the implementation of FCIT* is available at \url{https://robotic-esp.com/code/fcitstar/}.

\bibliography{refs}{}
\bibliographystyle{ieeetr}

\end{document}